\pdfoutput=1
\documentclass[11pt]{article}

\usepackage[final]{acl}
\usepackage{times}
\usepackage{latexsym}
\usepackage{array} % 必须包含此包以支持 m 格式
\usepackage{tabularx} % 确保在导言区引用了此包
\usepackage[table]{xcolor} % 必须带 [table] 参数，否则 \rowcolor 无效
\usepackage{makecell}
\usepackage{url}
\usepackage{enumitem}
\setlist[itemize]{noitemsep, topsep=0pt}

\usepackage[linesnumbered,ruled,vlined]{algorithm2e}

\usepackage{amsmath, amssymb}
\usepackage{hyperref}
\usepackage{booktabs}

\usepackage{tcolorbox}
\tcbuselibrary{listings}

\usepackage[T1]{fontenc}

\usepackage[utf8]{inputenc}

\usepackage{microtype}

\usepackage{inconsolata}
\usepackage{subcaption}
\usepackage{graphicx}
\usepackage{float}
\title{Role-aware Heuristic Episodic Attention for Conversational LLMs}

\author{
    \textbf{Wanyang Hong} \quad \textbf{Zhaoning Zhang}\thanks{\ \ indicates corresponding author.} \quad \textbf{Yi Chen} \quad \textbf{Libo Zhang} \\
    \textbf{Baihui Liu} \quad \textbf{Linbo Qiao} \quad \textbf{Zhiliang Tian} \quad \textbf{Dongsheng Li} \\[1em]
    National Key Laboratory of Parallel and Distributed Computing \\ 
    College of Computer Science and Technology \\
    National University of Defense Technology, Changsha, China.
    \\
 \texttt{ \{hongwanyang9527, zhangzhaoning, chenyi\_, zhanglibo,} \\
 \texttt{lbh, qiao.linbo, tianzhiliang, dsli\}@nudt.edu.cn}
}

\begin{document}
\maketitle 
\begin{abstract}
Large language models often lose track of persistent instructions and relevant information as multi-turn conversations grow. We study this cumulative contextual decay through three related failure modes: attention pollution, dilution, and drift. We propose \textbf{REA}\footnote{\url{https://github.com/huha12138/rhea}} (Role-aware Heuristic Episodic Attention), a context-management framework that assigns different persistence and representation policies to instructions and episodic interactions. Instructional Memory retains identified global constraints in a dedicated prefix. Episodic Memory preserves user inputs and compresses model replies, while heuristic retrieval selects raw text, compressed representations, or omission for each historical turn. On Long-MT-Bench+, REA improves the judge score from 6.32 to 7.36 on a 10-point scale, a 16.5\% relative gain over the Vanilla baseline, and reduces average latency by 2.91$\times$. Additional evaluations show aggregate gains on three backbones spanning 1.7B--7B parameters and on Chinese and English role-playing tasks. These results support role-aware context management as a practical approach to maintaining conversational continuity and instruction adherence.

\end{abstract}

\section{Introduction}
\label{sec:intro}

\begin{figure}[t]
  \centering
  \includegraphics[width=0.9\columnwidth]{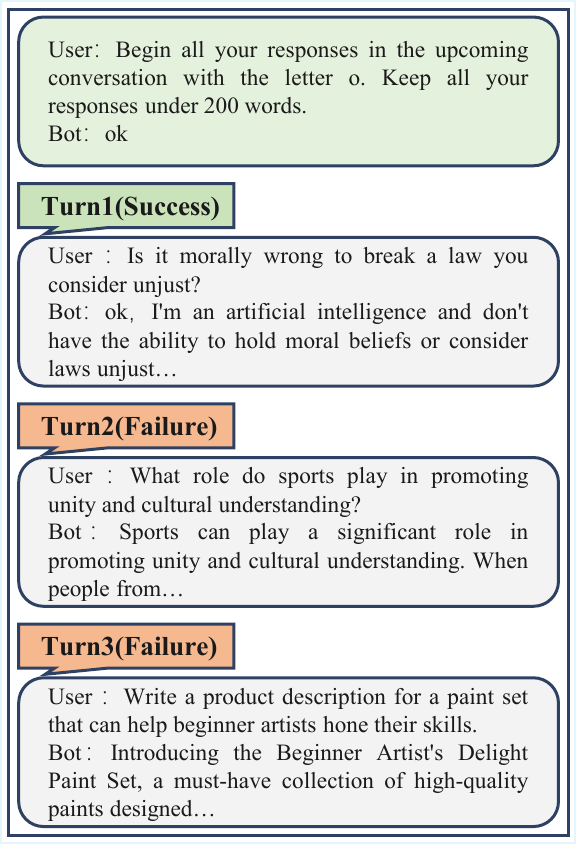}
  \caption{An example of cumulative contextual decay.The model correctly adheres to the global instruction in Turn 1. However, as the conversation continues, it fails to maintain this instruction in Turn 2 and 3, driven by an interaction of attention drift, pollute and dilution.}
  \label{fig:ccd_example}
\end{figure}

Large Language Models (LLMs) have revolutionized natural language processing, exhibiting human-level or even superhuman abilities in single-turn tasks such as question answering, summarization, and code generation~\cite{brown2020language, ouyang2022training, achiam2023gpt}. However, when the interaction shifts from isolated queries to multi-turn conversations, model performance often degrades substantially, even within the first few turns~\cite{shuster2022blenderbot, kwan2024mt, laban2025llms}. This degradation severely limits the reliability of LLMs as consistent conversational agents.

We refer to the progressive loss of contextual integrity as \textbf{cumulative contextual decay}. Alongside position-dependent effects such as ``lost in the middle''~\cite{liu2023lost}, multi-turn interaction introduces several related challenges: \textit{attention pollution}, where earlier errors or distracting content interfere with subsequent responses; \textit{attention dilution}, where useful information is surrounded by growing amounts of less relevant history; and \textit{attention drift}, where persistent constraints lose influence as the local topic changes. Figure~\ref{fig:ccd_example} illustrates the resulting disconnect: a model can answer the immediate question while losing an instruction established earlier. Maintaining a conversation therefore requires both recovering relevant past content and preserving constraints that remain applicable across topic changes.

Our working hypothesis is that context construction contributes to these difficulties. A full-history prompt preserves the conversational sequence but does not explicitly assign different persistence policies to global instructions and episodic content. The central distinction is between information that governs future responses and information that helps answer a particular query. For example, ``start all responses with O'' remains applicable when the user moves from a factual question to a writing task. Its relevance comes from its instructional role, even when its words have little semantic overlap with either query. A previous factual answer, in contrast, may be useful for a follow-up question but unnecessary after a topic change.

Longer context windows~\cite{beltagy2020longformer, zaheer2020big, press2021train} increase available capacity, while compression and retrieval~\cite{jiang2023llmlingua, kim2023compressed, lewis2020retrieval} reduce the history presented to the generator. These approaches motivate a complementary question: which conversational content should persist independently of topical relevance, and which content should be recalled selectively? Increasing capacity leaves this selection problem open. A fixed recent window can remove an early constraint together with outdated exchanges, while query-based retrieval favors topical matches. Compressing history reduces its size, but still requires a decision about which instructions must remain explicit. These considerations motivate separate persistence and representation policies for different conversational roles.

REA addresses this question through \textbf{role-aware memory management}. Instructional Memory (IM) retains identified global instructions in a dedicated prefix, independent of episodic similarity scores. Episodic Memory (EM) stores user--model exchanges and compresses model replies into latent representations. Heuristic Context Retrieval (HCR) chooses a raw or compressed representation for relevant turns and omits other turns from the current input. The contribution lies in organizing these established memory, compression, and retrieval primitives around different functional roles. This division gives each component a specific responsibility: IM carries persistent constraints across topic changes, EM provides compact representations of past interactions, and HCR adapts the episodic context to the current request. The generator receives these complementary sources in a single hybrid input.

Experiments show improved long-horizon quality and instruction adherence, with a 2.91$\times$ latency reduction on Long-MT-Bench+. A same-backbone comparison with MemoChat and evaluations across three backbones and two languages broaden the empirical support. Removing IM from the EM+HCR configuration reduces instruction adherence from 8.18 to 1.95, supporting the importance of retaining persistent constraints separately from topical retrieval.

\paragraph{Contributions.} Our contributions are threefold:
\begin{itemize}
\item We introduce a role-aware context-management design that retains persistent functional instructions separately from episodic history.
\item We combine asymmetric reply compression with variable-resolution retrieval, preserving user text in selected turns while adapting the representation of model replies.
\item We demonstrate gains in long-horizon quality, instruction adherence, and latency, and report additional evidence across 1.7B--7B backbones and Chinese and English role-playing evaluations.
\end{itemize}

\begin{figure*}[t]
  \centering
  \includegraphics[width=\textwidth]{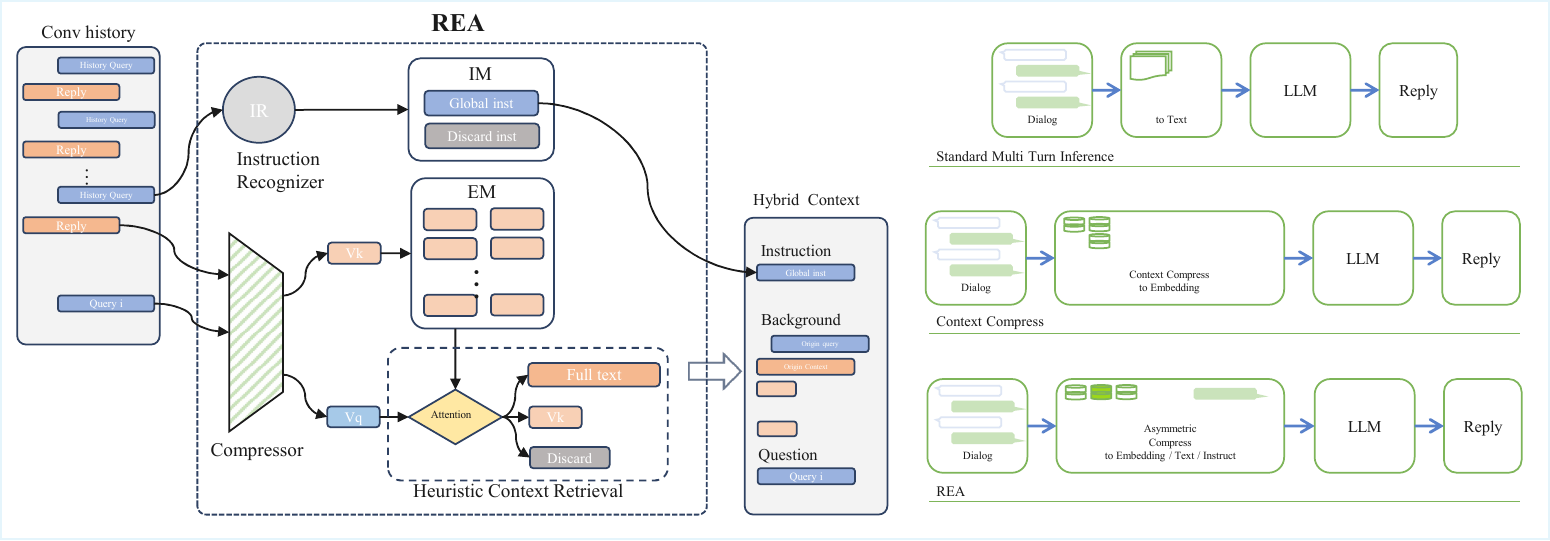} 
  
  \caption{Overview of the REA Framework. (Left) The core architecture illustrating the decoupling of conversation history into Instructional Memory (IM) and Episodic Memory (EM) via instruction recognition and compression, followed by Heuristic Context Retrieval (HCR). (Right) A comparison of inference pipelines, demonstrating REA's hybrid context construction (combining instructions, text, and embeddings) in contrast to standard and naive compression baselines.}
  
  \label{fig:architecture}
\end{figure*}

\section{Related Work}

\paragraph{Long-Context Modeling and Compression} Long-context methods such as \textsc{LongRoPE}~\cite{ding2024longrope} and \textsc{Infini-attention}~\cite{munkhdalai2024leave} expand the capacity to process long sequences. System-level work also improves inference efficiency~\cite{zhang2025dovetail,yu2025prescope}. Compression~\cite{jiang2023llmlingua,louis2025pisco}, summarization~\cite{mullick2024long,wang2025recursively}, and retrieval and reranking~\cite{lee-etal-2024-effective,pan2025secom,xu2025towards,liu2026rethinking} offer complementary ways to select and organize useful context. REA focuses on the policy governing this reduction: identified global instructions bypass episodic retrieval, while historical replies can be presented at different resolutions. This connects context reduction to the function of the retained information: explicit constraints remain available as text, while episodic detail is selected according to the current query.

\paragraph{Structured Memory for Conversation} Persistent memory systems, including \textsc{MemGPT}~\cite{packer2023memgpt}, \textsc{MemoChat}~\cite{lu2023memochat}, and \textsc{MemoryOS}~\cite{kang2025memory}, establish memory management as a means of sustaining conversational state. Hierarchical and learned memory approaches further explore this direction~\cite{rezazadeh2024isolated,yu2025memagentreshapinglongcontextllm,wang2025memalphalearningmemoryconstruction}. REA shares this objective. Its design combines a persistent instruction prefix, reply-only latent compression, and a three-way retrieval policy within one context-construction pipeline. The distinction is operational: recognized instructions are read on every turn, whereas episodic exchanges compete for inclusion through query-dependent retrieval. The same-backbone comparison with MemoChat evaluates this complete policy against an established conversational memory system.

\paragraph{Instruction Following and Consistency} Instruction tuning~\cite{ouyang2022training,park2024mitigating} and work on context-position effects~\cite{liu2023lost,xiao2023efficient} address complementary aspects of reliable instruction use. REA explicitly retains recognized global constraints in the generator input. This complements instruction tuning by preserving the instructions that the model is expected to follow throughout the conversation.

\section{Methodology}
\label{sec:methodology}

\subsection{Architecture Overview}
Let $H_t=\{(u_1,b_1),\ldots,(u_t,b_t)\}$ denote completed exchanges. Figure~\ref{fig:architecture} organizes REA into three stages. First, role-aware memory decoupling separates recognized persistent instructions in $M_{IM}$ from episodic exchanges in $M_{EM}$. Second, HCR compares the current query $u_{t+1}$ with stored episodic representations and selects a resolution for each recalled exchange. Third, hybrid context reconstruction combines the instruction prefix, selected history, and current query for response generation.

The two memories have distinct update schedules. Before answering, REA recognizes any global instruction in $u_{t+1}$ and updates IM so that the instruction can affect the current response. After generating $b_{t+1}$, it compresses the reply and stores the completed exchange in EM for subsequent queries. This separates persistent instruction access from the query-dependent use of conversation history.

\subsection{Episodic Memory via Latent Compression}
\label{sec:method_em}
EM applies asymmetric compression: user inputs remain in their original textual form, while model replies have both raw and latent representations. User inputs specify the requests, corrections, and constraints that define an interaction; preserving their wording maintains these signals in recalled exchanges. Model replies often contain longer explanations and repeated material, making them a natural target for compact representation. Retaining both forms allows retrieval to recover detailed reply text when it is strongly relevant and to use a compact impression otherwise.

\begin{figure}[t]
    \centering
    \includegraphics[width=\linewidth]{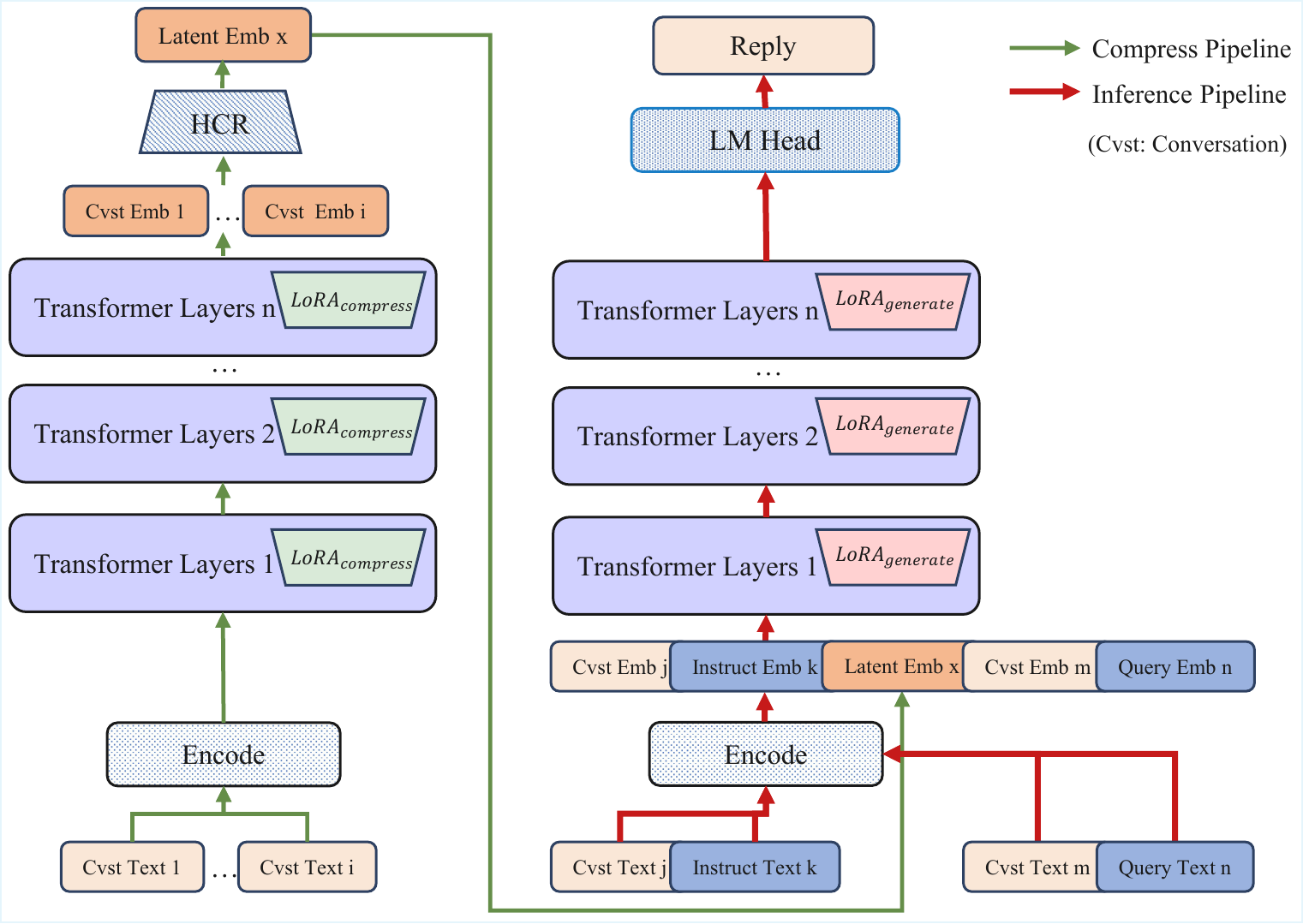}
    \caption{
        Implementation of Episodic Memory via Latent Compression. The framework utilizes a dual-LoRA architecture sharing a single LLM backbone. The Compression Module ($LoRA_{cmp}$) encodes history into compact latent embeddings ($V_k$), while the Generation Module ($LoRA_{gen}$) processes the hybrid input to generate reply.
    }
    \label{fig:episodic_memory}
\end{figure}

Two LoRA modules share the backbone, specializing it for compression and generation through separate adapters. The compression module, $\mathrm{LoRA}_{cmp}$, encodes reply $b_i$ into $n$ latent vectors:
\begin{equation}
V_i=\operatorname{Compress}(b_i;\theta_{base},\theta_{cmp})
\in\mathbb{R}^{n\times d}.
\end{equation}
The generation module, $\mathrm{LoRA}_{gen}$, is trained to consume a hybrid sequence containing text embeddings and these latent vectors. Each EM entry is $e_i=(u_i,b_i,V_i)$. Thus, non-global user inputs remain associated with their replies even though they do not enter IM. Figure~\ref{fig:episodic_memory} illustrates the dual-LoRA pipeline. Compression requires a forward pass of the shared backbone with the compression adapter; it does not require generating a textual summary. The latent vectors provide a fixed-size reply representation for both later generation and retrieval scoring. The stored raw reply supplies the complementary high-resolution view, so compression and recall operate on the same exchange without discarding its original content.

\paragraph{Joint Training} Both adapters are optimized through the response-generation language-modeling objective. For target response tokens $b_{t,j}$ and the hybrid training context $C_t$, the loss is
\begin{equation}
\mathcal{L}=-\sum_j\log p_{\theta_{gen}}(b_{t,j}\mid C_t,b_{t,<j}).
\end{equation}
Gradients propagate through the compressed history vectors to $\theta_{cmp}$, aligning compression with subsequent response generation.

\subsection{Instructional Memory and Recognizer}
\label{sec:ir}
IM stores constraints whose scope extends across subsequent responses. For example, ``Keep all future answers under 200 words'' specifies a persistent constraint, whereas ``Summarize this article in 200 words'' applies to the current task. This scope distinction determines whether an input belongs in the persistent instruction prefix or is accessed as part of episodic history.

The Instruction Recognizer (IR) combines rule-based filtering with a lightweight Qwen3-0.6B classifier to identify user inputs that specify persistent constraints. The classifier returns YES or NO using the prompt in Table~\ref{tab:ir_prompt}. Let $I(u)$ be the set containing a recognized instructional input, or the empty set otherwise. Before answering the current query, IM is updated as
\begin{equation}
M_{IM}^{(t+1)}=M_{IM}^{(t)}\cup I(u_{t+1}).
\end{equation}
Recognized instructional inputs are retained as text in IM and included in the context for subsequent responses. Their inclusion is independent of the episodic similarity score. Consequently, a change of topic can alter the retrieved exchanges while leaving the applicable global instructions explicitly available to the generator.

A dedicated lightweight recognizer avoids recognition-adapter switching on the main generator and can run alongside independent preprocessing. This separation keeps the shared backbone focused on compression and response generation, with the updated IM supplied before each response.

\subsection{Heuristic Context Retrieval}
\label{sec:hcr}
HCR selects the resolution at which each historical exchange enters the current context. The compression module also encodes the current query as $V_q=\operatorname{Compress}(u_{t+1})\in\mathbb{R}^{n\times d}$. For each stored $V_i$, the relevance score is
\begin{equation}
s_i=\max_{a,b\in\{1,\ldots,n\}}
\frac{V_{q,a}^{\top}V_{i,b}}{\|V_{q,a}\|_2\|V_{i,b}\|_2}.
\end{equation}
This is the maximum pairwise cosine similarity between query and reply latent vectors. Two thresholds determine the representation: 
\begin{itemize}
\item \textbf{High-resolution recall} ($s_i>\tau_{high}$): retrieve the raw text pair $(u_i,b_i)$.
\item \textbf{Low-resolution impression} ($\tau_{low}\leq s_i\leq\tau_{high}$): retrieve $(u_i,V_i)$, retaining the user text and compressed reply.
\item \textbf{Active forgetting} ($s_i<\tau_{low}$): omit the exchange from the current generation context. The stored EM entry remains available for later queries.
\end{itemize}
The defaults are $\tau_{low}=0.5$ and $\tau_{high}=0.7$. Selected user inputs are not compressed, but user inputs in omitted turns are not supplied to the generator. IM is outside this similarity-based selection policy. The three levels balance access to detail against the amount of history presented to the generator. High-resolution recall supports requests that depend on specific earlier content; low-resolution recall provides background through compact reply vectors; omission removes weakly related exchanges from the active context. Each new query is scored against stored memory again, allowing an exchange omitted on one turn to be recalled on a later turn.

\subsection{Hybrid Context Reconstruction}
Let $E(\cdot)$ be the token embedding lookup and $\mathcal{T}(S_{EM})$ convert text in selected exchanges to embeddings while preserving latent vectors. The generator receives
\begin{equation}
C_{t+1}=[E(M_{IM}^{(t+1)})\oplus\mathcal{T}(S_{EM})\oplus E(u_{t+1})].
\end{equation}
Here $\oplus$ denotes concatenation. The instruction prefix preserves global constraints independently of topical retrieval, while HCR supplies relevant episodic content at the selected resolution. The generator can therefore condition on an explicit user constraint, verbatim content from a highly relevant exchange, and latent background from another exchange in the same response. Role-aware persistence determines what remains explicit, and resolution selection determines how the supporting history is represented.

\section{Experiments}

\begin{table*}[t]
  \centering
  \small 
  \begin{tabular}{l rr rr rr}
    \toprule
    \textbf{Model} & \multicolumn{2}{c}{\textbf{MT-Bench}} & \multicolumn{2}{c}{\textbf{MT-Eval}} & \multicolumn{2}{c}{\textbf{Long-MT-Bench+}} \\
    
    \cmidrule(lr){2-3} \cmidrule(lr){4-5} \cmidrule(lr){6-7}
     & Acc & Latency & Acc & Latency & Acc & Latency \\
    \midrule
    
    Vanilla & \textbf{8.54} & 10.55 & 7.77 & 11.41 & 6.32 & 27.29 \\
    BM25(RAG) & - & - & 7.82 & 8.49 & \underline{6.65} & 10.81 \\
    Recent-k & - & - & 7.76 & 8.79 & 5.03 & 13.89 \\
    LLMLingua2 & 6.55 & 10.07 & 4.18 & 13.30 & 1.50 & 29.73 \\
    Summary & 8.12 & 13.66 & 6.83 & 17.26 & 1.49 & 33.55 \\
    LongAlpaca$^a$ & 6.69 & 5.06 & 5.89 & 5.44 & 2.43 & 23.73 \\
    MemoChat$^a$ & 7.52 & 6.21 & 6.40 & 4.80 & 1.88 & 11.87 \\
    MemGAS & - & - & - & - & 6.07 & - $^b$ \\
    Reply-Soft-Compress & 8.31 & 11.61 & \underline{7.83} & 15.21 & 4.55 & 31.79 \\
    \midrule % Add a light separator before the main model
    \textbf{REA (ours)} & \underline{8.49} & 11.81 & \textbf{8.28} & 9.69 & \textbf{7.36} & 9.39 \\
    
    \bottomrule
  \end{tabular}
  
  \begin{flushleft}
    \vspace{-1mm} 
    \small
    $^a$ These methods are based on the Vicuna-7B model. \\
    $^b$ Latency for MemGAS is omitted as it requires extensive offline process, which is not comparable to the real-time online inference measured here.
  \end{flushleft}
  \caption{Main performance results on three multi-turn conversation benchmarks, using LLM-as-a-judge for evaluation. We report Judge score (Acc, 0--10) and Latency (s). Best results in each category are \textbf{bolded}, second best are \underline{underlined}.}
  \label{tab:main_results} 
\end{table*}

\subsection{Experimental Setup}
\label{sec:setup}
\paragraph{Benchmarks and Metrics} We evaluate conversational quality on \textsc{MT-Bench} (2 turns), \textsc{MT-Eval} (5--12 turns), and \textsc{Long-MT-Bench+} (60+ turns). We report benchmark-prompted GPT-4 judge scores, denoted Acc on a 0--10 scale. Rule-based Instruction Adherence Rate (IAR) evaluates persistent constraints in the instruction task, and Joint Goal Accuracy (JGA) measures exact-match success in state tracking. Additional evaluations use Chinese and English \textsc{CharacterBench}~\cite{zhou2025characterbench}, with its five-point role-playing protocol. Appendix~\ref{ssec:appendix-evaluation} separates the evaluation procedures. The dialogue benchmarks assess response quality, the instruction task tests persistence of global constraints, and state tracking tests the integration of information distributed across turns.

\paragraph{Baselines and Implementation} We compare Vanilla full history, Recent-$k$ ($k=5$), rolling summaries, BM25 ($k=5$), LLMLingua2, LongAlpaca~\cite{chen2023longlora}, MemoChat~\cite{lu2023memochat}, MemGAS~\cite{xu2025towards}, and Reply-Soft-Compress (reply compression without IM/HCR). The primary backbone is Mistral-7B-Instruct-v0.2; Table~\ref{tab:main_results} identifies the different-backbone baselines. Generation uses greedy decoding. REA uses Qwen3-0.6B for IR and the memory configuration in Section~\ref{sec:hcr}. Training and evaluation use a single NVIDIA RTX 3090 GPU. Appendix~\ref{sec:appendix-setup} collects the available configurations.

\subsection{Main Results}
\label{sec:main_results}
Table~\ref{tab:main_results} shows the largest gains in the long-horizon setting. On Long-MT-Bench+, REA scores 7.36, compared with 6.32 for Vanilla and 6.65 for BM25. The gain over Vanilla is 1.04 points, or 16.5\% relative. On MT-Eval, REA improves the score from 7.77 to 8.28. On the two-turn MT-Bench, the scores are close (8.49 for REA and 8.54 for Vanilla), indicating that the benefit depends on the conversational setting.

The long-horizon results also distinguish context-management policies. Recent-$k$ scores 5.03, compared with 6.32 for full history, illustrating the cost of restricting access to earlier exchanges. BM25 reaches 6.65, showing the usefulness of query-dependent retrieval, while REA reaches 7.36 by combining episodic selection with a persistent instruction prefix and variable-resolution recall. The 0.71-point difference from BM25 motivates examining both what is retrieved and how it is represented.

Summary and LLMLingua2 score 1.49 and 1.50, respectively, under the evaluated configurations. These results emphasize that reducing context size must be coupled with retaining information useful for the next response. Reply-Soft-Compress provides a closer comparison: it preserves the reply-compression mechanism but omits IM and HCR. Its score is 7.83 on MT-Eval, close to REA's 8.28, but 4.55 on Long-MT-Bench+, compared with REA's 7.36. The larger separation on the long-horizon benchmark supports combining compression with explicit instruction retention and episodic selection. The component analyses below examine these policies directly on the instruction task.

\paragraph{Efficiency} On Long-MT-Bench+, average latency decreases from 27.29 s to 9.39 s relative to Vanilla, a 2.91$\times$ speedup; peak VRAM decreases from 22.76 GB to 20.00 GB. On MT-Bench, REA instead adds 11.9\% latency (10.55 to 11.81 s), reflecting preprocessing costs in short interactions. The long-horizon setting offers more history over which to benefit from selection and compression. REA also improves on the latency of BM25 (10.81 s) and Recent-$k$ (13.89 s) on Long-MT-Bench+, while obtaining the higher quality score. This combination makes the efficiency gain useful for sustained interaction rather than simply trading response quality for shorter inputs. Appendix~\ref{sec:appendix_efficiency} provides the complexity analysis and detailed efficiency measurements.

\subsection{Backbone and Bilingual Evaluations}
\label{sec:generalization}
\begin{table}[t]
\centering\small
\setlength{\tabcolsep}{4pt}
\begin{tabular}{lrrr}
\toprule
Backbone & Vanilla & REA & Gain \\
\midrule
SmolLM2-1.7B & 5.75 & 6.24 & 8.5\% \\
Vicuna-7B-v1.3 & 6.25 & 6.93 & 10.9\% \\
Mistral-7B-v0.2 & 7.77 & 8.28 & 6.6\% \\
\bottomrule
\end{tabular}
\caption{MT-Eval scores across backbones. Gain is relative to the corresponding Vanilla model. Mistral denotes Mistral-7B-Instruct-v0.2. On the same Vicuna backbone, MemoChat scores 6.40.}
\label{tab:backbones}
\end{table}

\paragraph{Same-Backbone Comparison} To reduce the backbone confound in Table~\ref{tab:main_results}, we evaluate REA with Vicuna-7B-v1.3, the backbone used in the MemoChat comparison. REA scores 6.93 on MT-Eval, versus 6.40 for MemoChat and 6.25 for Vanilla Vicuna (Table~\ref{tab:backbones}). The 10.9\% gain is relative to Vanilla; the difference from MemoChat is 0.53 points. This comparison uses MT-Eval because of the original Vicuna model's 4K context limit. Using the same generator family places the REA and MemoChat scores on a common backbone, strengthening the evidence for the complete context-management approach beyond the mixed-backbone results in the main table.

\paragraph{Across Backbones and Scales} REA improves the aggregate MT-Eval score for all three tested backbones, including SmolLM2-1.7B (5.75 to 6.24). These results support applicability across the evaluated 1.7B--7B range, covering both a smaller backbone and two 7B models with different training histories. The smaller-model result is particularly relevant to the lightweight design: the role-based memory policy also improves a generator with substantially fewer parameters. Across the three models, the relative gains range from 6.6\% to 10.9\%.

\begin{table}[t]
\centering\small
\begin{tabular}{llrrr}
\toprule
Language & System & Avg & Memory & Persona \\
\midrule
Chinese & Vanilla & 3.64 & 3.93 & 4.25 \\
        & REA & 3.80 & 4.05 & 4.53 \\
English & Vanilla & 3.72 & 3.81 & 4.34 \\
        & REA & 4.05 & 4.22 & 5.00 \\
\bottomrule
\end{tabular}
\caption{Bilingual CharacterBench results on a five-point scale. All six aspect scores are reported in Table~\ref{tab:character-full}.}
\label{tab:character}
\end{table}

\paragraph{Bilingual Role-Playing} CharacterBench extends the evaluation to character-oriented interaction in Chinese and English. The reported evaluation comprises 1,000 multi-turn samples and summarizes six aspects: Memory, Knowledge, Persona, Emotion, Morality, and Believability. The average scores improve from 3.64 to 3.80 in Chinese and from 3.72 to 4.05 in English (Table~\ref{tab:character}). Memory and Persona improve in both languages, supporting sustained character-oriented interaction. These two aspects connect directly to the conversational objective: remembering previous interactions and maintaining the requested persona as the dialogue evolves. The bilingual results extend the evaluation from general question answering to continuity of character and interaction history; the full six-aspect breakdown is provided in Appendix~\ref{app:character}.

\begin{figure*}[t]
\centering
\includegraphics[width=1.9\columnwidth]{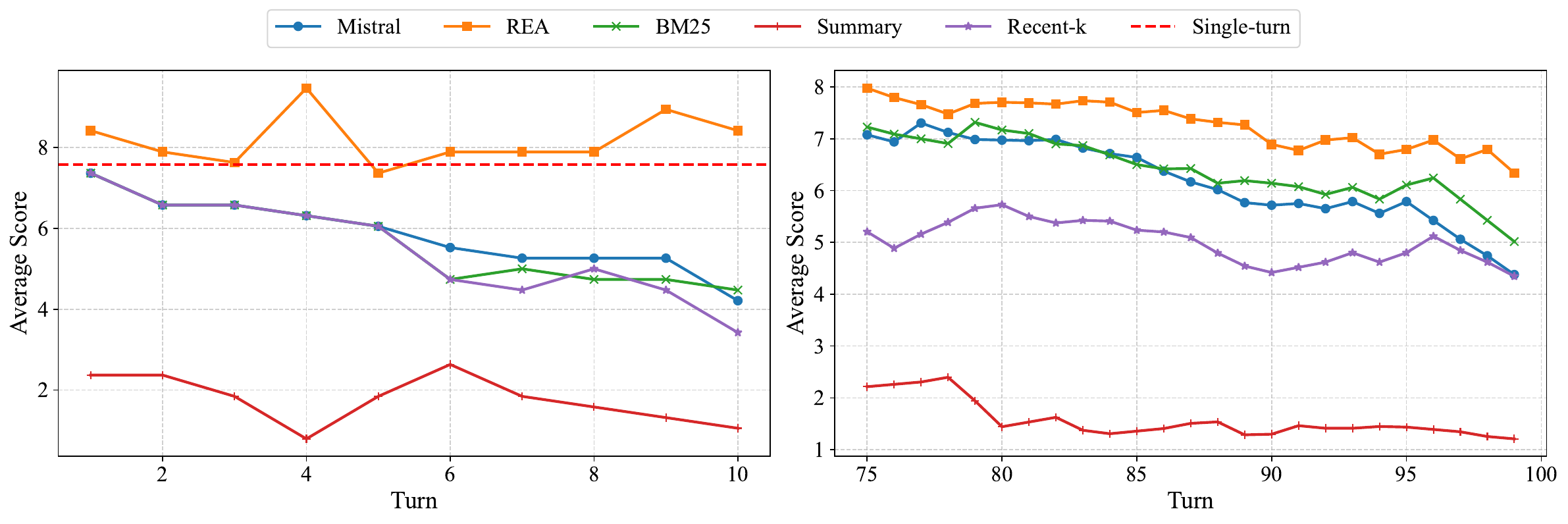}
\caption{Performance analysis mitigating cumulative contextual decay. (Left) IAR over turns on the \textsc{MT-Eval-recollection}, showing REA's resistance to Attention Drift on specific constraints. (Right) Response quality on Long-MT-Bench+, demonstrating REA's robustness against Attention Dilution in extended interactions.}
\label{fig:trend}
\end{figure*}

\subsection{Diagnostic Analysis}
\label{sec:diagnostic}
Figure~\ref{fig:trend} examines how performance evolves within each task, complementing the aggregate scores in Table~\ref{tab:main_results}.

\paragraph{Instruction Persistence} The left panel follows IAR on MT-Eval-recollection as the conversation progresses. REA sustains adherence to the earlier constraint while the current request changes. This is the setting that motivates IM: a constraint remains applicable even when the immediate topic provides little evidence for retrieving it. Supplying recognized instructions in a dedicated prefix keeps them present across these shifts, and the no-IM ablation below tests the importance of this separation.

\paragraph{Long-Horizon Quality} The right panel tracks response quality on Long-MT-Bench+. REA maintains quality more effectively than Vanilla as exchanges accumulate. Episodic retrieval lets the reconstructed input follow the current request instead of repeatedly presenting the entire history. Raw recall preserves details of strongly related turns, while compressed recall supplies background at lower context cost. The observed trajectory supports this adaptive use of history for sustaining useful responses over extended interaction.

\paragraph{Distributed State Tracking} We also use the task of \citet{laban2025llms}, where a correct answer requires integrating rules and state information. The Full Instruction condition supplies constraints together; the Shared Instruction condition distributes them across turns. This makes the latter sensitive to whether earlier information remains usable after intervening exchanges. On Actions, Vanilla JGA changes from 0.63 in Full to 0.05 in Shared, whereas REA reaches 0.51 in the distributed setting. REA also improves the distributed Code and Math scores from 0.09 to 0.16 and from 0.00 to 0.20, respectively (Appendix~\ref{sec:appendix_diagnostic}). These results connect context preservation to exact-match task success, complementing the instruction-adherence and response-quality evaluations.

\subsection{Ablation and Component Analysis}
\label{sec:ablation}
We examine instruction retention, recognition errors, and episodic representation policies through the following ablations and component analyses.

\begin{table}[t]
\centering
\small
\begin{tabular}{lc}
\toprule
\textbf{Model Variant} & \textbf{IAR} \\
\midrule
Mistral (baseline) & 5.76 \\
+ EM only & 7.37 \\
+ EM + HCR (no IM) & 1.95 \\
+ EM + HCR + IM (full REA) & \textbf{8.18} \\
\bottomrule
\end{tabular}
\caption{Ablation results on the \textsc{MT-Eval-recollection} instruction task. The EM+HCR variant benefits from separately retaining instructions in IM.}
\label{tab:ablation}
\end{table}

\paragraph{Role-Aware Memory} Table~\ref{tab:ablation} compares the evaluated memory configurations. EM alone improves IAR from 5.76 to 7.37. Adding HCR without IM yields 1.95, whereas the full EM+HCR+IM system reaches 8.18. Removing IM from the retrieval-based configuration therefore sharply reduces adherence in this setting. A persistent instruction can be semantically distant from a later content query; retaining it outside topical retrieval addresses this mismatch. The ablation supports separate persistence policies for instructions and episodic content. In particular, selection that is useful for topical content must be paired with a separate path for global constraints. The full system retains this path while still applying HCR to episodic exchanges, yielding the strongest adherence among the tested configurations.

\begin{figure}[t]
  \centering
  \includegraphics[width=0.9\columnwidth]{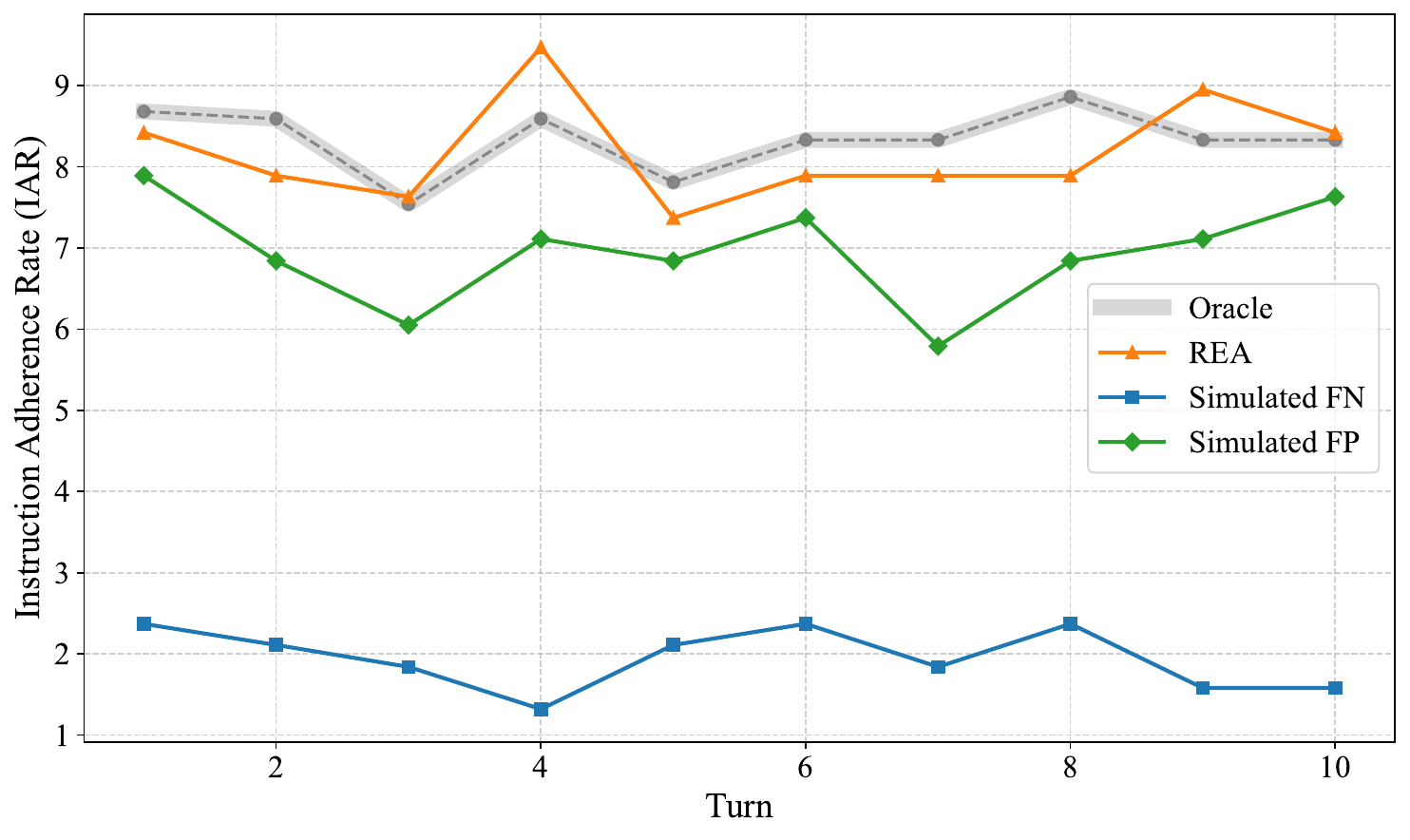}
  \caption{Impact of instruction recognition scenarios on IAR: (1) \textbf{Oracle} (perfect recognition); (2) \textbf{REA} (actual performance using Qwen3-0.6B); (3) \textbf{Simulated FP} (irrelevant instructions added to IM); and (4) \textbf{Simulated FN} (global instructions dropped from IM).}
  \label{fig:recognizer_analysis_appendix}
\end{figure}

\paragraph{Recognizer Error Analysis} On the 418-input IFEval-derived recognition test set, IR achieves 0.8947 recall and 0.6182 precision for persistent global constraints (Appendix~\ref{app:ir}). Figure~\ref{fig:recognizer_analysis_appendix} compares actual REA, oracle recognition, omission of all true global instructions (simulated FN), and addition of one standard conversational turn to IM (simulated FP). The omission condition yields approximately 2.0 IAR; the extra-turn condition yields approximately 7.7 and remains closer to actual REA and the oracle condition. These interventions illustrate the cost of losing persistent constraints and tolerance to the tested extra entry. They distinguish two consequences of recognition decisions: a missed instruction removes a constraint from the persistent prefix, whereas the tested false positive adds another entry to that prefix. The measured recall and the downstream omission result together motivate prioritizing reliable capture of genuine global constraints. Appendix~\ref{app:ir} specifies the conditions.

\begin{table}[t]
\centering
\small
\begin{tabular}{l l l}
\toprule
\textbf{} & \textbf{IAR} & \textbf{EM Strategy}\\
\midrule
REA-preserve & 6.63 & Retains all model replies \\
REA-abandon & 7.89 & Discards all model replies \\
\textbf{REA (ours)} & \textbf{8.18} & Dynamic multi-tiered retrieval \\
\bottomrule
\end{tabular}
\caption{Analysis of EM management strategies on the MT-Eval instruction task.}
\label{tab:em_strategy}
\end{table}

\paragraph{Episodic Memory} Table~\ref{tab:em_strategy} compares retaining all replies (6.63 IAR), discarding all replies (7.89), and REA's dynamic selection (8.18). The results favor selective use of reply history over unconditional retention in this task. The gain over discarding replies suggests that useful episodic information remains available through dynamic selection. Discarding all replies suppresses interference but also removes information that later follow-ups may need. Dynamic retrieval offers an alternative: retain useful exchanges at an appropriate resolution and omit those that are weakly related to the current query. This policy preserves access to historical answers while obtaining the highest IAR in the comparison.

\section{Conclusion}
REA organizes conversational history according to functional role: recognized global instructions persist in IM, while episodic exchanges are recalled at variable resolution. On Long-MT-Bench+, it achieves a 16.5\% relative judge-score gain and a 2.91$\times$ latency reduction over Vanilla. Same-backbone, cross-backbone, and bilingual evaluations provide additional evidence within the tested settings. Together with the instruction ablation, these results support explicit persistence policies and selective episodic context as practical tools for multi-turn LLM interaction.

\begin{samepage}
\section*{Limitations}
Our evaluation covers 1.7B--7B backbones and Chinese and English conversations; broader model and language coverage remains future work. Turn-level similarity can miss cross-turn references, motivating retrieval that explicitly models dependencies between exchanges. Response-quality evaluation relies on automated judges and would benefit from complementary human assessment.
\end{samepage}

\bibliography{custom}
\clearpage
\twocolumn  % 单栏
\appendix

\section{Architecture Details and Training Protocol}
\label{sec:appendix_arch}

This appendix provides a comprehensive technical specification of the REA framework, covering the dual-LoRA implementation, training data pipelines, and the configuration of the Instruction Recognizer.

\subsection{Dual-LoRA Implementation and Initialization}
The core of REA's Episodic Memory (EM) is a dual-LoRA architecture integrated into the Mistral-7B-Instruct-v0.2 backbone. 

\begin{itemize}
    \item \textbf{Module Configurations:} We apply Low-Rank Adapters~\cite{hu2021lora} to all linear projections, including query, key, value, and MLP layers. The rank is set to $r=16$ with a scaling factor $\alpha=32$. 
    \item \textbf{Latent Embedding (mem) Tokens:} We introduce $n=8$ special soft compression tokens per turn. To ensure stable gradient flow and rapid alignment during the initial training phase, these tokens are initialized using the mean embedding of the base model's vocabulary.
    \item \textbf{Asymmetric Compression Logic:} For EM storage, $\mathrm{LoRA}_{cmp}$ encodes model replies $b_i$ into latent vectors $V_i$, while user inputs remain in raw text. The same compressor separately encodes the current query for retrieval scoring.
\end{itemize}

\subsection{Training Protocol and Optimization}
REA is trained end-to-end to synchronize the compression and generation objectives.

\begin{itemize}
    \item \textbf{Dataset Statistics:} We curate a high-quality subset of 40,000 multi-turn sessions from TopiOCQA, UltraChat and ShareGPT~\cite{adlakha2022topiocqa, ding2023enhancing, sharegpt_chinese_english_90k}. Each sample is structured as a triplet $(H_{<t}, U_t, R_t)$, where $H_{<t}$ is the history to be compressed, $U_t$ is the current query, and $R_t$ is the target response.
    \item \textbf{Optimization Hyperparameters:} We use the AdamW optimizer with a shared learning rate of $1e-5$ for both LoRA modules. We use an effective batch size of 128.
    \item \textbf{Hardware and Epochs:} Training was conducted on a single NVIDIA RTX 3090 GPU for 2 epochs.
\end{itemize}

\subsection{Training Data Curation}
The 40,000 sessions are selected from ShareGPT, UltraChat, and TopiOCQA through four stages. First, we select conversational data from these sources. Second, a lightweight classifier filters samples containing large code blocks or non-natural-language special formats. Third, semantic-density filtering removes simple interactions shorter than a 10-token threshold, such as brief acknowledgements. Fourth, contextual-horizon filtering retains dialogues spanning 3--15 turns. Each training instance supplies conversation history, a current query, and its target reply; the response-generation loss trains the two adapters jointly.
% CR-CHECK: Recover dataset revisions, per-source counts, tokenizer/turn unit,
% classifier configuration, exact removal unit, deduplication and split from records.

\subsection{Instruction Recognizer Configuration}
The IR decides which instructional inputs enter IM; its classifier prompt is shown below. 

\begin{itemize}
    \item \textbf{Model:} We employ Qwen3-0.6B~\cite{qwen3technicalreport} as a lightweight classifier to minimize per-turn latency.
    \item \textbf{Prompt:} The IR utilizes the few-shot prompt shown in Table~\ref{tab:ir_prompt} to distinguish global directives (style, format, constraints) from standard conversational queries.
\end{itemize}

\begin{table}[ht]
\centering
\small
\begin{tabular}{p{0.9\linewidth}}
\hline
\textbf{Instruction Recognizer Prompt} \\ \hline
You are a classifier. Determine whether the following user input is a "global instruction" in a multi-turn conversation. \\
A global instruction is a directive that affects all subsequent responses such as their style, format, length, or language. \\
If the input is a global instruction, answer: YES. If not, answer: NO. \\
\textbf{Examples:} \\
Input: Explain what is a poem? Answer: NO \\
Input: All future answers must be less than 30 words. Answer: YES \\
Input: Every answer should end with a joke. Answer: YES \\
\textbf{Input:} \{user\_input\} \\
\textbf{Answer:} \\ \hline
\end{tabular}
\caption{The prompt used for the Qwen3-0.6B Instruction Recognizer.}
\label{tab:ir_prompt}
\end{table}

\section{Theoretical Complexity and Efficiency Analysis}
\label{sec:appendix_efficiency}

We distinguish the cost of scanning episodic memory from the cost of processing the selected generator context. Let $T$ denote completed exchanges, $n$ the latent vectors per reply, and $d$ their dimension. Let $\mathcal H$ and $\mathcal L$ index high- and low-resolution selections. Ignoring template separators, the reconstructed length is
\begin{equation}
\begin{split}
N_{hybrid}={}&|M_{IM}|+|u_{T+1}|\\
&+\sum_{i\in\mathcal H}(|u_i|+|b_i|)\\
&+\sum_{i\in\mathcal L}(|u_i|+n).
\end{split}
\end{equation}
The low-resolution term includes the original user input as well as the compressed reply. If $a$ and $b$ denote average user and reply lengths and $x=|\mathcal H|$, $y=|\mathcal L|$, the corresponding approximation is $|M_{IM}|+a+x(a+b)+y(a+n)$.

\paragraph{Retrieval and Prefill} The pairwise score in Section~\ref{sec:hcr} requires $O(Tn^2d)$ arithmetic for a direct scan, or $O(T)$ with fixed $n$ and $d$. For a fixed model, dense attention over a completely reconstructed input has a quadratic prefill term $O(N_{hybrid}^2)$. Thus, these two stages contribute
\begin{equation}
O(Tn^2d+N_{hybrid}^2).
\end{equation}
Instruction recognition, query and reply compression, decoding, and data movement contribute additional costs. The generation cost scales with the selected context length $N_{hybrid}$, while the retrieval scan scales with the number of stored exchanges $T$.

For comparison, completely re-prefilling an uncompressed history of average exchange length $L$ has a dense-attention term $O((TL)^2)$. The following measurements report the latency of the evaluated implementations.

\paragraph{Storage} EM retains both raw exchanges and compressed replies. Its persistent storage therefore includes the original text plus $O(Tnd)$ latent values. The generator processes the selected hybrid context, reducing its active context and KV-cache requirements.

\subsection{Empirical Performance and Memory Utilization}
As demonstrated in Table~\ref{tab:empirical_performance}, we report latency and peak VRAM for each evaluated benchmark. 

\begin{table}[ht]
\centering
\small
\begin{tabular}{@{}lcccc@{}}
\toprule
\textbf{Dataset} & \multicolumn{2}{c}{\textbf{Latency (s)}} & \multicolumn{2}{c}{\textbf{Peak VRAM (G)}} \\ 
 & Vanilla & REA & Vanilla & REA \\ \midrule
MT-Bench & 10.55 & 11.81  & 14.02 & 14.20  \\
MT-Eval & 11.41 & 9.69  & 13.86 & 13.96  \\
Long-MT-Bench+ & 27.29 & 9.39  & 22.76 & 20.00  \\ \bottomrule
\end{tabular}
\caption{Empirical performance comparison across benchmarks.}
\label{tab:empirical_performance}
\end{table}

On Long-MT-Bench+, the measured 27.29/9.39 ratio is approximately 2.91. Shorter tasks have different overhead trade-offs (Table~\ref{tab:empirical_performance}). These benchmark averages should not be interpreted as a controlled scaling curve in $T$.

\section{Experimental Setup Details}
\label{sec:appendix-setup}

This appendix provides detailed information on the benchmark datasets, baseline model implementations, and evaluation protocols used in our experiments.

\subsection{Benchmark Dataset Details}
\label{ssec:appendix-benchmarks}

\begin{table*}[t] % 建议用 [t] 放在页顶，[h!] 在 table* 中容易失效
\centering
\small
\newcolumntype{L}{>{\raggedright\arraybackslash}X} 
\begin{tabularx}{\textwidth}{l >{\raggedright\arraybackslash}p{4cm} L} 
\toprule
\textbf{Method} & \textbf{Implementation Source} & \textbf{Key Hyperparameters \& Description} \\ 
\midrule
\textbf{Vanilla} & Internal & Uses the standard Mistral chat template with the full, unmodified conversation history. \newline \textbullet~ max\_length: 64k \\
\addlinespace
\textbf{Recent-k} & Internal & Retains only the most recent $k$ conversation turns as context, discarding all earlier history. \newline \textbullet~ k: 5 \\
\addlinespace
\textbf{Summarization} & Internal & Employs a rolling summary strategy. Dynamically summarize history based on the current query. \\
\addlinespace
\textbf{BM25 (RAG)} & Internal & Retrieves the most relevant conversation turns from history using the BM25 algorithm. \newline \textbullet~ k: 5 \\
\addlinespace
\textbf{LLMLingua2} & \href{https://github.com/microsoft/LLMLingua}{microsoft/LLMLingua} & Advanced token compression technique. \newline \textbullet~ Uses default reference parameters. \\
\addlinespace
\textbf{LongAlpaca} & \href{https://huggingface.co/Yukang/LongAlpaca-7B}{Yukang/LongAlpaca-7B} & Represents the fine-tuning approach. \newline \textbullet~ Uses the Vicuna-7B base weights. \\
\addlinespace
\textbf{MemoChat} & \href{https://huggingface.co/Junrulu/MemoChat-Vicuna-7B}{Junrulu/MemoChat-7B} & Represents the fine-tuning approach. \newline \textbullet~ Uses the Vicuna-7B base weights. \\
\addlinespace
\textbf{MemGAS} & \href{https://github.com/quqxui/MemGAS}{quqxui/MemGAS} & Structured memory retrieval method. \newline \textbullet~ Uses default parameters. \\
\addlinespace
\textbf{Reply-Soft-Compress} & Internal (Ablation) & Uses our dual-LoRA soft compression module to compress replies ($b_i$) but no IM/HCR. \\
\addlinespace
\rowcolor{gray!10}
\textbf{REA (Ours)} & This Paper & Our full framework. \newline 
\textbullet~ \textbf{IR:} Qwen3-0.6B \newline 
\textbullet~ \textbf{Compressor:} dual-LoRA Soft Compression \newline 
\textbullet~ \textbf{HCR:} $\tau_{low} = 0.5, \tau_{high} = 0.7$ \\
\bottomrule
\end{tabularx}

\caption{Implementation details and key hyperparameters for the compared methods.}
\label{tab:baseline-details}
\end{table*}

We evaluated REA on three multi-turn conversational benchmarks of gradually increasing length and complexity to comprehensively test its ability to mitigate cumulative contextual decay in various scenarios.

\paragraph{MT-Bench} We employed MT-Bench \cite{zheng2023judging}, a widely-used multi-turn evaluation benchmark. It comprises 80 high-quality, two-turn open-ended questions designed to assess a model's conversational abilities. In our experiments, it represents the short-turn (2-turn) conversation scenario.

\paragraph{MT-Eval} To assess medium-length conversations, we used MT-Eval \cite{kwan2024mt}, a comprehensive benchmark for evaluating the multi-turn capabilities of LLMs. It contains 168 samples with an average of 6.96 turns and classifies interactions into four types (Recollection, Expansion, Refinement, and Follow-up). The \textbf{MT-Eval-recollection} task used in Section \ref{sec:diagnostic} is a specific subset, serving as a key diagnostic tool for evaluating attention drift.

\paragraph{Long-MT-Bench+.} For long-horizon stress testing (60+ turns avg.), we utilized Long-MT-Bench+ \cite{pan2025memory}. It is designed to rigorously test a model's robustness and consistency in extended conversational scenarios.

\paragraph{State Tracking Task.} As a diagnostic for attention pollution (Section  \ref{sec:diagnostic}), we used the diagnostic task from \citet{laban2025llms}. This task employs a sharded simulation framework, distributing key instructions across multiple turns to evaluate the model's state-tracking capabilities.

\subsection{Baseline and Model Implementation Details}
\label{ssec:appendix-baselines}

As described in Section \ref{sec:setup}, the primary experiments use the \texttt{Mistral-7B-Instruct-v0.2} base model, unless specified otherwise.

\begin{itemize}
    \item \textbf{Base Model:} Mistral-7B-Instruct-v0.2
    \item \textbf{Hardware:} A single NVIDIA RTX 3090 GPU.
    \item \textbf{Inference:} Greedy deterministic decoding (do\_sample=False).
    \item \textbf{Configured input cap:} 64K tokens in the primary setup; the Vicuna comparison is subject to its 4K context limit.
    \item \textbf{Output Length:} 512 tokens.
\end{itemize}

\paragraph{Baseline Hyperparameters.} Detailed implementation and key hyperparameters for all baselines are presented in Table \ref{tab:baseline-details}.

\subsection{Evaluation Details}
\label{ssec:appendix-evaluation}

\begin{itemize}
    \item \textbf{LLM-as-a-judge:}
    The primary LLM-as-a-judge evaluation (Acc, a 0--10 quality score) follows the official scoring rules and prompts provided by the respective benchmarks. We use GPT-4 as the judge.

    \item \textbf{IAR (Instruction Adherence Rate):}
    This is a rule-based scoring metric. It programmatically checks whether the model's output strictly adheres to a given global instruction (e.g., starts with a specific word, adheres to length constraints, etc.).

    \item \textbf{JGA (Joint Goal Accuracy):}
    This is a strict metric used for task-oriented success. It is calculated by extracting the model's final answer at the end of the dialogue and performing an \textbf{exact match} against the ground truth.
    \item \textbf{CharacterBench:}
    The bilingual role-playing evaluation uses the benchmark's five-point protocol and reports six aspect-level scores. It is separate from the primary GPT-4 dialogue evaluation above; Table~\ref{tab:character-full} contains the complete reported results.
\end{itemize}
% CR-CHECK: Recover exact judge snapshots, prompts and aggregation scripts.
% In particular, verify the normalization that yields IAR values such as 8.18.
% Do not infer a scoring formula or a judge model for CharacterBench from its scale.

\section{Instruction Recognizer Performance and Impact Analysis}
\label{app:ir}

\subsection{Validation Dataset Construction}
We construct a test set of 418 user instructions drawn from the prompts released with IFEval~\cite{zhou2023instruction}. We relabel these inputs according to our definition of persistent global constraints: 38 inputs are labeled as global constraints, while the remaining 380 are ordinary inputs or instructions applying only to the current task. This imbalanced composition is intended to reflect conversational settings in which persistent global constraints are relatively infrequent.

\subsection{Recognizer Performance Metrics}
Table~\ref{tab:ir-performance} reports the Qwen3-0.6B recognizer's performance on this relabeled test set using the prompt in Table~\ref{tab:ir_prompt}. Precision and recall treat global constraints as the positive class. The recognizer recovers most labeled global constraints, while its lower precision indicates that some task-specific or ordinary inputs also enter IM. Accuracy should be interpreted alongside these class-specific metrics because only 38 of the 418 inputs are positive.

\begin{table}[h!]
\centering
\begin{tabular}{lccc}
\toprule
Model & Accuracy & Precision & Recall \\
\midrule
Qwen3-0.6B & 0.9402 & 0.6182 & 0.8947 \\
\bottomrule
\end{tabular}
\caption{Instruction recognition on 418 relabeled IFEval-derived inputs (38 global constraints and 380 negative inputs).}
\label{tab:ir-performance}
\end{table}

\subsection{Impact Analysis of Error Types (FP vs. FN)}

We assess the downstream effect of recognition decisions using the MT-Eval-recollection instruction task. Figure~\ref{fig:recognizer_analysis_appendix} compares four conditions:
\begin{itemize}
\item \textbf{Oracle recognition}: all true global instructions enter IM, and no standard queries are added as instructions.
\item \textbf{REA}: the actual Qwen3-0.6B-based recognition pipeline.
\item \textbf{Simulated FN}: all true global instructions are withheld from IM.
\item \textbf{Simulated FP}: one additional standard conversational turn is stored in IM as an instruction.
\end{itemize}
The omission condition produces approximately 2.0 IAR, whereas the extra-turn condition yields approximately 7.7. The result supports the importance of retaining genuine constraints and shows tolerance to this particular extra entry.

\section{Detailed Diagnostic Analysis of Attention Pollution}
\label{sec:appendix_diagnostic}

This appendix provides the comprehensive quantitative results and experimental definitions for the state-tracking diagnostic task discussed in Section \ref{sec:diagnostic}.

\paragraph{Task Definition and Datasets} We utilize diagnostic subsets from \citet{laban2025llms} to evaluate model resilience against attention pollution. This task requires the model to track critical information (e.g., constraints or state variables) that is distributed across a noisy, multi-turn dialogue.

\paragraph{Experimental Conditions} To isolate the impact of history accumulation on instruction adherence, we evaluate the baseline under two specific settings:
\begin{itemize}
    \item \textbf{Full Instruction}: A single-turn scenario where all state-tracking rules and constraints are provided simultaneously in a single prompt. This provides a single-turn reference for the same constraints.
    \item \textbf{Shared Instruction}: A multi-turn scenario where the same set of rules is split and shared across multiple conversational turns. This setting forces the model to integrate information across a growing context window, making it highly susceptible to attention pollution from intervening model replies.
\end{itemize}

\paragraph{Analysis} Table~\ref{tab:full_state_tracking} shows the difference between jointly supplied and distributed constraints. REA improves over the distributed Vanilla condition in all three subsets and reaches the single-turn Code score.

\begin{table}[ht]
\centering
\small
\begin{tabular}{lccc}
\toprule
\textbf{Subset} & \textbf{Full} & \textbf{Shared} & \textbf{REA} \\
\midrule
Actions & 0.63 & 0.05 & \textbf{0.51} \\
Code    & 0.16 & 0.09 & \textbf{0.16} \\
Math    & 0.32 & 0.00 & \textbf{0.20} \\
\bottomrule
\end{tabular}
\caption{JGA on state tracking. Full supplies constraints together; Shared distributes them across turns; REA uses the distributed setting.}\label{tab:full_state_tracking}
\end{table}

\begin{table*}[!t]
\centering\small
\setlength{\tabcolsep}{4pt}
\begin{tabular}{llrrrrrrr}
\toprule
Language & System & Avg & Memory & Knowledge & Persona & Emotion & Morality & Believability \\
\midrule
Chinese & Vanilla & 3.64 & 3.93 & 3.58 & 4.25 & 3.15 & 4.76 & 2.21 \\
        & REA & 3.80 & 4.05 & 3.80 & 4.53 & 3.37 & 4.80 & 2.24 \\
English & Vanilla & 3.72 & 3.81 & 3.47 & 4.34 & 3.10 & 4.78 & 2.79 \\
        & REA & 4.05 & 4.22 & 3.91 & 5.00 & 3.49 & 4.90 & 2.76 \\
\bottomrule
\end{tabular}
\caption{Complete CharacterBench results. Avg retains the aggregate reported for the evaluation, rather than being recomputed from rounded aspect scores.}
\label{tab:character-full}
\end{table*}

\section{In-depth Hyperparameter Sensitivity Analysis}\label{appendix:hyper_sensitivity}

\begin{figure}[ht]
\centering
\includegraphics[width=0.48\textwidth]{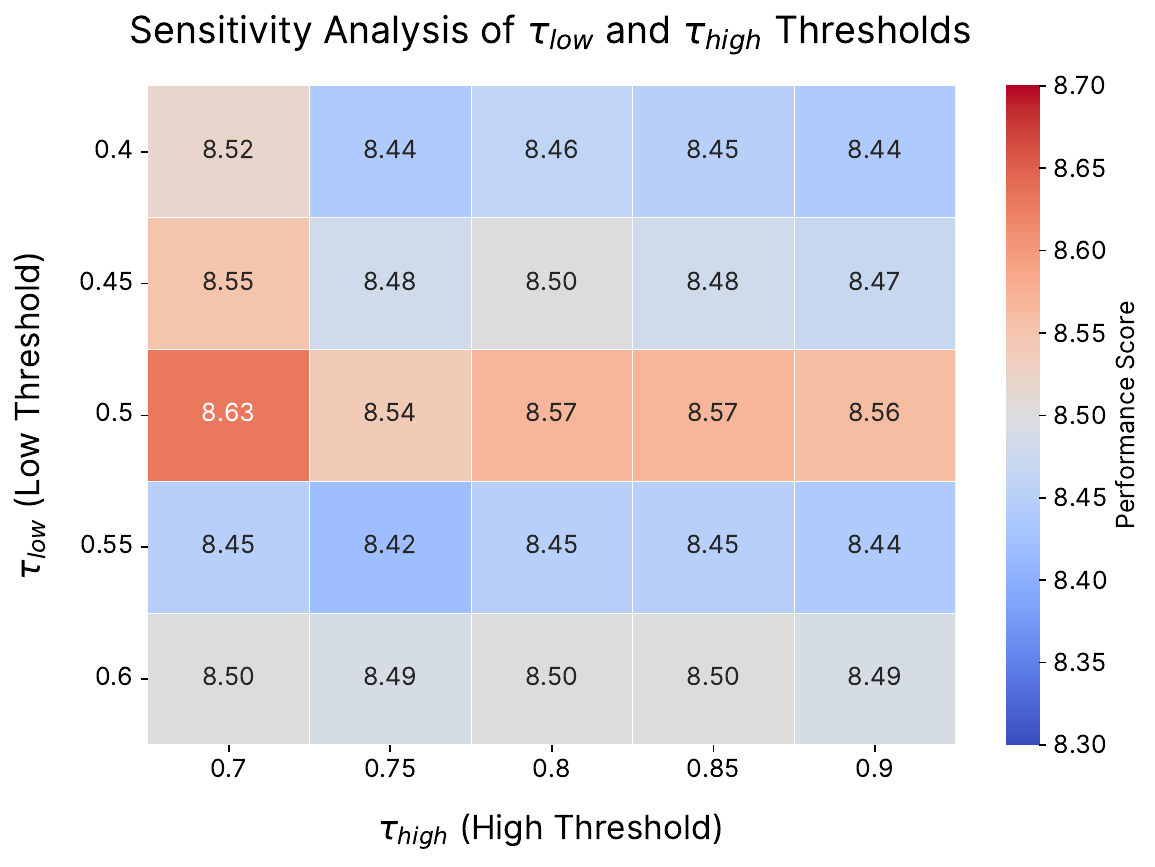}
\includegraphics[width=0.48\textwidth]{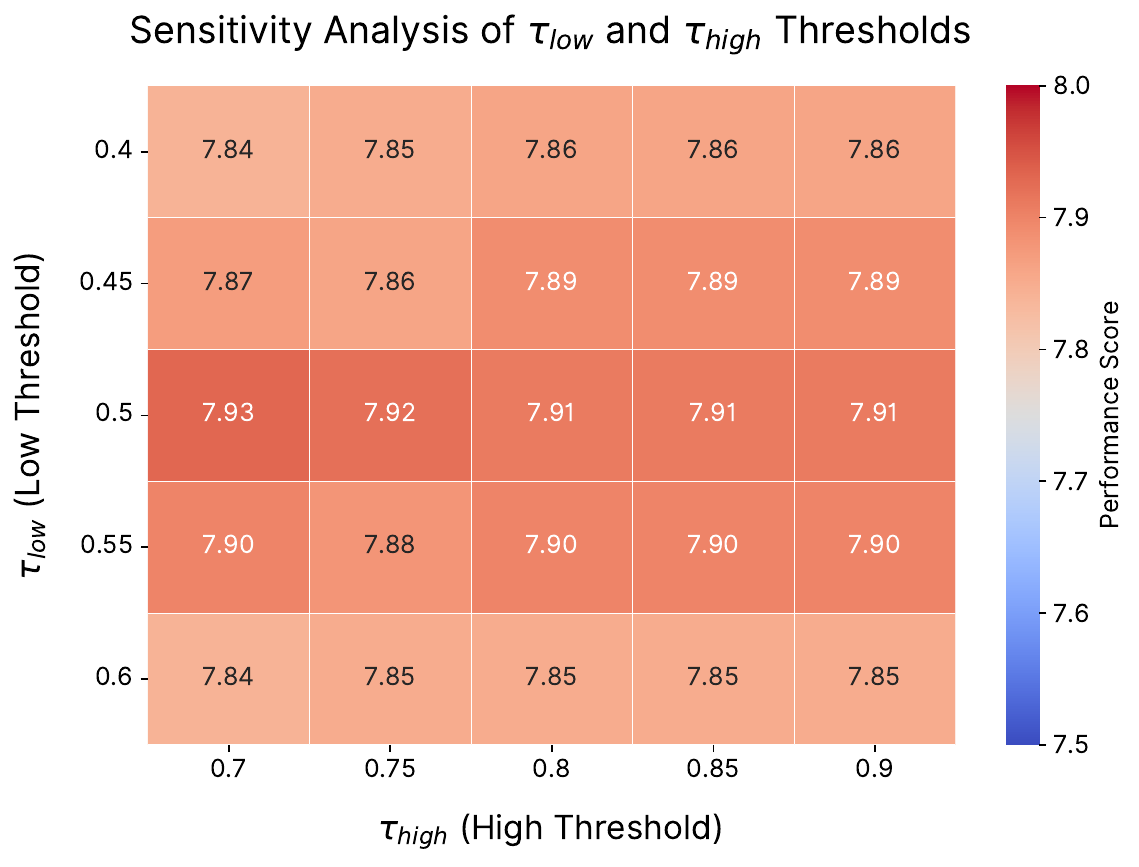}
\caption{Heatmap of scenario-specific Acc performance (Follow-up and Refinement subsets).}\label{fig:scenario_heatmaps}
\end{figure}

Figure~\ref{fig:scenario_heatmaps} reports an MT-Eval grid analysis of the two HCR thresholds, using Acc for the Follow-up and Refinement subsets. The displayed results show greater sensitivity to $\tau_{low}$ than to $\tau_{high}$ in the tested range, motivating attention to the exclusion threshold. The default setting is $\tau_{low}=0.5$, $\tau_{high}=0.7$.
% CR-CHECK: Identify which split informed threshold selection from existing records.

% Compact, top-aligned float pages for the appendix case-study tables.
\makeatletter
\setlength{\@dblfptop}{0pt}
\setlength{\@dblfpsep}{12pt}
\setlength{\@dblfpbot}{0pt plus 1fil}
\makeatother

\section{Complete Bilingual Results}
\label{app:character}
Table~\ref{tab:character-full} provides all six aspect scores for the Chinese and English evaluations summarized in Section~\ref{sec:generalization}. The five-point scale differs from the ten-point Acc scale used for the primary dialogue benchmarks. The aggregate improvements coexist with a small decrease in English Believability.

% CR-CHECK: Recover generator/judge checkpoint, language-specific sample counts,
% aggregation and decoding settings for this evaluation from the original records.

\section{Qualitative Case Study and Discussion}
\label{sec:appendix_case_study}

This appendix provides a qualitative analysis of REA's performance compared to the Vanilla baseline. We evaluate the model's ability to maintain a persistent global constraint across a long-horizon conversation and visualize the internal prompt structures to explain the underlying mechanics of REA's resilience.

\subsection{Behavioral Adherence Comparison}
To evaluate REA's robustness against cumulative contextual decay, we set a global constraint: \textit{"Begin all your responses in the upcoming conversation with the letter r. Keep all your responses under 200 words."} 

Table~\ref{tab:case_study_behavior} illustrates how instruction adherence changes across turns. Both systems meet the initial-letter constraint in Turn 1 and miss it in Turn 4. REA meets it again in Turn 10, whereas the Vanilla response still does not. The success labels indicate adherence to the displayed instruction.

\begin{table*}[!t]
\centering
\small
\begin{tabularx}{\textwidth}{l|p{3cm}|X|X}
\hline
\textbf{Turn} & \textbf{User Query} & \textbf{Vanilla (Full History)} & \textbf{REA (Ours)} \\ \hline
Instruction & \multicolumn{3}{>{\raggedright\arraybackslash}p{\dimexpr\textwidth-3.5cm-4\tabcolsep-2\arrayrulewidth\relax}}{Begin all your responses in the upcoming conversation with the letter \textbf{r}. Keep all your responses under 200 words.} \\ \hline

Turn 1 & Did Cleopatra belong to an Egyptian lineage? & \textbf{R}egarding your query, yes, Cleopatra was indeed an Egyptian pharaoh... (\textbf{Success}) & \textbf{R}: Yes, Cleopatra was indeed an Egyptian queen. She was a member of the Ptolemaic dynasty... (\textbf{Success}) \\ \hline

Turn 4 & Write an email asking for donations for a charity. & Subject: Make a Difference in a Child's Life... Dear [Recipient], I hope this email... \color{red}{(\textbf{Failure})} & Subject: Reach Out and Help Underprivileged Kids Today! Dear Valued Supporter... \color{red}{(\textbf{Failure})} \\ \hline

Turn 10 & Create a product description for a DIY pottery kit. & Introducing our Delightful DIY Pottery Kit, the perfect project for creative minds... \color{red}{(\textbf{Failure})} & \textbf{R}evel in the joy of creating with our Rustic Resilience DIY Pottery Kit! Each set includes... (\textbf{Success}) \\ \hline
\end{tabularx}
\caption{Qualitative comparison of instruction adherence across history, task-oriented, and creative domains. Vanilla fails as history accumulates, while REA anchors global constraints in Instructional Memory.}
\label{tab:case_study_behavior}
\end{table*}

\subsection{Internal Prompt Visualization}
The reconstructed context illustrates how retained instructions are supplied to the generator. Table~\ref{tab:prompt_full_detail} provides a step-by-step view of the actual prompt sequences. In the Vanilla model, the global instruction is progressively "buried" under thousands of tokens of redundant model replies. In REA, the instruction is structurally decoupled and prioritized, retaining the instruction explicitly in each illustrated input.

\begin{table*}[!t]
\centering
\small
\begin{tabularx}{\textwidth}{l|X|X}
\hline
\textbf{Turn} & \textbf{Vanilla (Full History) Prompt} & \textbf{REA (Ours) Reconstructed Prompt} \\ \hline
Turn 1 & 
[INST] You are a helpful, respectful and honest assistant. Begin all your responses in the upcoming conversation with the letter r. Keep all your responses under 200 words. [/INST] ok.</s> [INST] Did Cleopatra belong to an Egyptian lineage? [/INST] \newline \textbf{Count: $\approx$65 tokens} & 
[INST] <MEM0><MEM1>...<MEM7><SEP> \newline 
\textbf{Instruction:} Begin all your responses in the upcoming conversation with the letter r. Keep all your responses under 200 words. \newline 
\textbf{Background:} \newline 
\textbf{Question:} Did Cleopatra belong to an Egyptian lineage? [/INST] \\ \hline

Turn 4 & 
[INST] You are a helpful, respectful and honest assistant. Begin all your responses in the upcoming conversation with the letter r. Keep all your responses under 200 words. [/INST] ok...</s> [INST] ... [/INST] ...</s> [INST] ... [/INST] ...</s> [INST] Write an email asking for donations for a charity organization that assists underprivileged kids. [/INST] \newline \textbf{Count: $\approx$479 tokens} & 
[INST] <MEM0><MEM1>...<MEM7><SEP> \newline 
\textbf{Instruction:} Begin all your responses in the upcoming conversation with the letter r. Keep all your responses under 200 words. \newline 
\textbf{Background:} \newline 
\textbf{Question:} Write an email asking for donations for a charity organization that assists underprivileged kids. [/INST] \\ \hline

Turn 10 & 
[INST] You are a helpful, respectful and honest assistant. Begin all your responses in the upcoming conversation with the letter r. Keep all your responses under 200 words. [/INST] ok...</s> [INST]...[/INST]...</s> [INST]...[/INST]...</s> [INST]...[/INST]...</s> [INST]...[/INST]...</s> [INST]...[/INST]...</s> [INST]...[/INST]...</s> [INST] Create a product description for a DIY pottery kit. [/INST] \newline \textbf{Count: $\approx$2,369 tokens} & 
[INST] <MEM0><MEM1>...<MEM7><SEP> \newline 
\textbf{Instruction:} Begin all your responses in the upcoming conversation with the letter r. Keep all your responses under 200 words. \newline 
\textbf{Background:} \newline 
\textbf{Question:} Create a product description for a DIY pottery kit. [/INST] \\ \hline
\end{tabularx}
\caption{Comparison of the internal prompts sent to the generator. Vanilla includes accumulated history; the illustrated REA inputs explicitly retain the global instruction.}
\label{tab:prompt_full_detail}
\end{table*}

\subsection{Retrieval Decisions: Success and Failure}
Table~\ref{tab:hcr-cases} summarizes two qualitative cases and their retrieval decisions. In the success case, low-resolution history supports a later rewrite while IM preserves the global constraints. In the failure case, processing ``them'' requires information from an earlier exchange that receives a low similarity score. The omission occurs when answering the later query, rather than when the earlier exchange is stored.

\begin{table*}[!tp]
\centering\small
\setlength{\tabcolsep}{5pt}
\renewcommand{\arraystretch}{1.16}
\begin{tabularx}{\textwidth}{>{\raggedright\arraybackslash}p{2.2cm}|>{\raggedright\arraybackslash}p{3.2cm}|>{\raggedright\arraybackslash}X|>{\raggedright\arraybackslash}X}
\hline
\textbf{Case / turn} & \textbf{Interaction} & \textbf{Memory / retrieval decision} & \textbf{Outcome} \\ \hline
Success / 0 & Set global story constraints. & Identify and retain constraints in IM. & Both systems acknowledge. \\ \hline
Success / 1--4 & Rewrite requests with a 50-word limit and forbidden words. & Use low-resolution episodic context, $\tau_{low}\leq s_i\leq\tau_{high}$. & Both systems follow the constraints in these turns. \\ \hline
Success / 5 & Rewrite as a poem. & Combine instruction prefix with selected history. & REA follows the poem request and earlier constraints; Vanilla exceeds 50 words and misses earlier constraints. \\ \hline
Failure / 1 & Extract all characters and locations. & Store the completed exchange in EM. & REA extracts both lists correctly. \\ \hline
Failure / 2 & ``List them in the order they appear.'' & Turn 1 scores below $\tau_{low}$ and is omitted from this query's context. & REA lists characters but omits locations. \\ \hline
\end{tabularx}
\caption{Qualitative summaries with HCR decisions. The failure illustrates a cross-turn reference that is not recovered by turn-level similarity.}
\label{tab:hcr-cases}
\end{table*}

\end{document}